\documentclass{article}
\usepackage{ijcai19}

\usepackage{times}
\usepackage{soul}
\usepackage{url}
\usepackage[utf8]{inputenc}
\usepackage[small]{caption}
\usepackage{graphicx}
\usepackage{amsmath}
\usepackage{booktabs}
 \usepackage{hyperref}
\usepackage{cleveref}
\usepackage{subfigure}

\title{One Network for Multi-Domains: Domain Adaptive Hashing with Intersectant Generative Adversarial Networks}

\author{
 Tao He$^{1}$ \and  
   Yuan-Fang Li$^1$\and
   Lianli Gao$^2$\and
   Dongxiang Zhang$^2$ \And
   Jingkuan Song$^{2}$\footnote{Contact Author}
      \affiliations
       $^1$Faculty of Information Technology,Monash University, 
Clayton, Australia.\\
       $^2$ Center for Future Media and School of Computer Science and Engineering,\\
University of Electronic Science and Technology of China, Chengdu, China.\\
      \emails
    \{tao.he,   
     yuanfang.li\}@monash.edu,
    \{lianli.gao,   dongxiang.zhang\}@uestc.edu.cn,
    jingkuan.song@gmail.com
}

\begin{document}

\maketitle

\begin{abstract}
\tolerance=1
\emergencystretch=\maxdimen
\hyphenpenalty=10000
\hbadness=10000
With the recent explosive increase of digital data, image recognition and retrieval  become a critical practical application. Hashing is an effective  solution to this problem, due to its low storage requirement and high query speed. However, most of past works focus on hashing in a single (source) domain. Thus, the learned hash function may not adapt well in a new (target) domain that has a large distributional difference with the source domain. In this paper, we explore an end-to-end domain adaptive learning framework that simultaneously and precisely generates discriminative hash codes and classifies target domain images. Our method encodes two domains images  into a semantic common space, followed by two independent generative adversarial networks   arming at crosswise reconstructing two domains' images,  reducing domain disparity and improving alignment in the shared space.
We evaluate our framework on {four} public benchmark datasets, all of which show that our method is superior to the other state-of-the-art methods on the tasks of object recognition and image retrieval.    

\end{abstract}

\section{Introduction}

 With the explosive increase of digital data, the efficient retrieval of images   in terms of time and storage  has become an increasingly important problem. Hashing-based techniques are a perfect approach to address this problem due to its high query speed and low storage cost. The main goal of hashing is to convert high-dimensional features into low-dimensional, discriminative and compact binary codes that preserve semantic information. Many efficient hash methods are based on deep neural networks. 
Most of them focus on single domain, such as  ITQ \cite{DBLP:journals/pami/GongLGP13}, BA \cite{DBLP:conf/cvpr/Carreira-Perpinan15}, QBH~\cite{DBLP:journals/pr/SongGLZS18}, video hash~\cite{DBLP:journals/tip/SongZLGWH18} and BDNN \cite{DBLP:conf/eccv/DoDC16}. As a result, such hash methods only exhibit good performance on datasets that have little distribution difference with the training data. 
Addressing this drawback, adaptive or cross-model hash techniques  \cite{Li_2018_CVPR} have been proposed in recent years to deal with the domain shift problem. These methods are trained on a \emph{source} dataset and then applied on a different \emph{target} dataset, which may have large distributional variances with the source domain.  In \cite{DBLP:conf/cvpr/VenkateswaraECP17}, the authors first proposed to use deep neural networks to learn representative hash codes, but they did not consider semantic information in the target domain and only applied a cross-entropy loss on the target domain. 
An end-to-end framework, DeDAHA \cite{Long:2018:DDA:3209978.3209999}, was proposed to learn domain adaptive hash codes by two loss functions that preserve semantic similarity in hash codes.
In \cite{Li_2018_CVPR}, the authors leveraged two adversarial networks to maximize the semantic information of the representations between different modalities, but their network uses the label information to bridge the domain gap between two modalities. 
Although these methods have achieved high performance, they typically operate under a supervised setting, assuming the availability of labeled data in the target domain. However, in the real world, access to labelled data for the target domain may be very limited or entirely unavailable. Moreover, it is usually assumed that the true labels of the target domain are unavailable, making adaptive  learning a more challenging problem.

Thus, more and more works focus on 
unsupervised domain adaption, like \cite{DBLP:conf/icml/XieZCC18}. The basic idea for domain adaptive learning is to embed the source and target domains into a common space so that both datasets in the latent layer have similar feature distributions. 
As for unsupervised domain adaption, one strategy is to train (under supervision) a classifier on the labeled source domain and then adapt it to a new domain. Some works \cite{Hu_2018_CVPR} focus on how to assign high-confidence pseudo labels to the target domain  and treat those predicted labels as the ground truth of the target domain to fine-tune the model. In  \cite{Zhang_2018_CVPR}, the authors  proposed a novel sample selection method aiming to select high-confidence labels. 
Another strategy is to project the source and target domain into a common space and reduce the domain disparity. Maximum Mean Discrepancy (MMD)   has been widely used as a measure of variance between distributions in a reproducing-kernel Hilbert space. Subsequently, many works use MMD to achieve nonlinear alignment of domains. Deep Domain Confusion \cite{DBLP:journals/corr/TzengHZSD14} leverages domain confusion loss to learn a representation which is semantically meaningful and domain invariant.

Recently, generative Adversarial Networks (GAN) \cite{NIPS2014_5423} was explored to generate new data  having the same distribution with the input data. Later on, many works started to study how to apply adversarial networks into adaptive learning.
 In~\cite{DBLP:conf/icml/GaninL15}, the authors adopted an adversarial training mechanism to address the domain shift problem. The key component is the training of a discriminator that judges whether the feature comes from the source domain or the target domain. Specifically, the feature learning component tries to fool the discriminator so that it cannot distinguish the origin of the features. When the discriminator cannot differentiate the origins, it means that the domain disparity has been reduced to a relatively low level. The main drawbacks of those methods are that the discriminators can only judge the overall domain's distribution but cannot distinguish whether their subspaces in the common space is invariant or aligned. Inspired by this, we concentrate on how to leverage the discriminator to distinguish whether those subspaces are aligned.

In this paper, we propose a new, unsupervised method to tackle the above deficiencies of existing cross-domain hash techniques. Our method simultaneously addresses the tasks of recognition and retrieval in a unified network. In summary, our main contributions are threefold.

\begin{itemize}
\item We develop a novel end-to-end transfer learning network which can not only learn the semantic hash codes for the unlabeled target domain but also predict labels of the new domain.

\item We employ two identical, but separate generative adversarial networks (GAN) to reduce source and target domain difference. Each of them independently generates data for both domains from the common space in which label information is preserved. 
 
\item We evaluate our method on four public benchmark datasets. Our results strongly demonstrate that our method outperforms the other  {state-of-the-art unsupervised methods}, Duplex, CDAN-M, I2I,  etc, in both object classification and image retrieval. 
\end{itemize}
 
\begin{figure*}[t]
    \centering
    \hspace{-1.5cm}
    \includegraphics[width=1.0\linewidth]{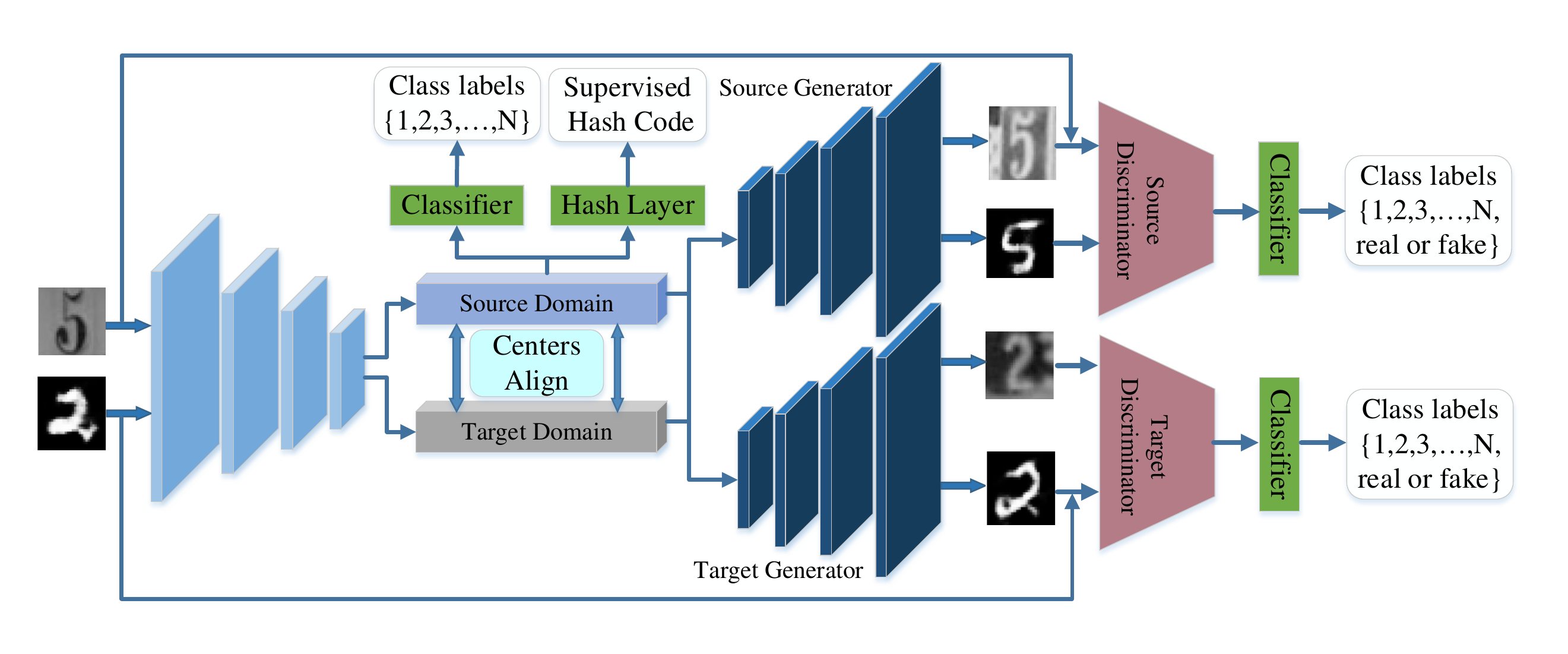}
    \hspace{-1.5cm}
    \caption{The overview of our framework, which is consisted of five networks: encoder, two independent generators and two distinct discriminators. Two generators are responsible for reconstructing two domain's data and discriminators aim at recogizing their labels.
      }
    \label{fig.framework}
\end{figure*}

\section{Methods} 

Let ${D_s} = \{(x_i^s,y_i^s)\}_{i = 1}^{{n_s}}$ denote the source domain, where $y_i^s \in \{ 1,\ldots,N\}$ is the \emph{label} of $x_i^s$. ${D_t} = \{ x_i^t\} _{i = 1}^{{n_t}}$ denotes the target domain. Note that the target domain $D_t$ is unlabeled. Our goal is to train a hash function in the source domain data and then test it in the target domain. 

The high-level architecture of our framework is shown in Figure \ref{fig.framework}. Our model consists of three components: (1) one shared encoder network $u=E({x})$, (2) two independent generators, denoted as $G^s$ for the source domain and $G^t$ for the target domain, and (3) two distinct discriminators $D^s$ and $D^t$, one for each domain. The shared encoder learns a common feature space for the two domains. Each generator generates images in both domains. Each discriminator judges whether an image comes from the source domain or the target domain. 

\subsection{Semantic Common Space Learning}

As shown in Figure \ref{fig.framework}, the semantic common space layer bridges the gap between the encoder and the generators. In this section, we will split our framework into three parts and illustrate them in detail.

\subsubsection{Supervised Hashing for Source Data}

Supervised hashing focuses on how to preserve the label similarity information into compact binary codes. 
In this work, we choose pairwise loss \cite{DBLP:conf/aaai/SongHGXHS18} as our supervised hashing function:
\begin{align}
     L_h = \mathop {\min }\limits_{{W^E}} (\frac{1}{2}{\sum\limits_{{s_{ij}} \in S^s} {(\frac{1}{d}{b_i}^T{b_j} - s_{ij})} ^2}  )
    \label{eq.sgn}
\end{align}
where $W^E$ is the set of parameters of the encoder, $b_i, b_j$ are binary codes, $S^s\in\{-1,1\}$ is the similarity matrix constructed from ground-truth labels of the source domain $D_s$, and $d $ is the length of hash code. Specifically, if two points have the same label, their similarity is defined as $1$ and otherwise $-1$. By optimizing Equation \ref{eq.sgn}, the hash function minimizes the feature distance of images with the same category but maximizes the distance across different categories. Unfortunately, hash codes $b_i, b_j$ are discrete and cannot be minimized directly. So we relax the hashing function into a continuous closed interval $[-1,1]$ and use ${u}_i$ to approximate the binary code $b_i$, where ${u}_i$ is the output of the last layer of network. We use $tanh(\cdot)$ as activation function to compute $u_i$. 
The updated loss function is given in Equation \ref{eq.sgn_}:
\begin{align}
 \label{eq.sgn_}
\begin{split}
L_h = &\mathop {\min }\limits_{{W^E}}( \frac{1}{2}{\sum\limits_{{s_{ij}} \in S^s} {(\frac{1}{d}u_i^T{u_j} - s_{ij})} ^2} + \\&
       \upsilon \frac{1}{2}{\sum {({u_i} - sign({u_i}))} ^2})
\end{split}
\end{align}
where the second term is the relaxation term aiming at reducing quantization error. In our experiments, we set a very small number for $\upsilon $.
 
\subsubsection{Semantic Centroid Alignment}

As noted in the introduction, the supervised hash loss learned in the source domain is inapplicable to the target domain in cross-domain hashing. To solve this problem, we proposed a \emph{semantic centroid alignment loss} to handle it. 
Especially, through our semantic centroid alignment loss, we force the target domain to have a cluster center distribution similar to that of the source domain. 

In fact, we can view Equation \ref{eq.sgn_} as a clustering process: if two images have the same class label, they should have a small hash distance (i.e.\     Hamming distance ), otherwise they should have a large hash distance. Consequently, a small hash distance means that their features should belong to the same cluster in the feature space whereas a large hash distance implies the they should belong in different clusters. Thus, if the two domains have similar cluster center distributions in the learned common semantic space, the hash function trained on the source domain should be applicable on the target domain. We adopt the K-means algorithm as the  clustering algorithm on the two domains. The formulation is showing as follows:
\begin{equation}
{L_s} = \mathop {\min }\limits_{{W^E}} (\sum\limits_{i = 1}^N {\varphi (\frac{1}{{{m_i}}}\sum\nolimits_{{y^s} = i} {{x^s}} ,\frac{1}{{{k_i}}}\sum\nolimits_{{y^{\tilde t}} = i} {{x^t}} )} )
\label{eqn:ls}
\end{equation}
where $N$ is the number of classes and $m_i$ (resp.\ $k_i$) denotes the number of samples in the same cluster in the source (resp.\ $k_i$) domain. $\varphi(\cdot,\cdot)$ is the function that measures the distance of different centers. In this work, we leverage Euclidean distance to define the distance between centers, i.e.\ $\varphi ({x_i},{x_j}) = {\left\| {{x_i} - {x_j}} \right\|^2}$. $y^{\tilde t}$ denotes the pseudo label of the target domain. To obtain highly precise pseudo labels, we set a threshold to select the target  label, as shown below:
\begin{align}
{y^{\tilde t}} = \left\{ \begin{array}{l}
\arg \max ({p}({x^t})){\rm{ }}~~~~~~\text{if~} {p}({x^t}) > T\\
 - 1{\rm{          }}~~~~~~~~~~~~~~~~~~~~~~~~~~~\text{otherwise}
\end{array} \right.
\label{eq.threshold}
\end{align}
where  $argmax(.)$ is the function to choose the index of the maximum in a vector and  $p(x^t)$ is the probability that $x^t$ belongs to each category and $T$ is the threshold.

\subsubsection{Label Prediction}

To predict pseudo labels, a classifier $C$ is learned on the common space layer $u$ with the following cross-entropy loss:
\begin{align}
{L_c} = \mathop {\min }\limits_{{W^E}} ( - \sum\limits_{i=1}^{{n_s}} {{y_i^s}\log (p({x_i^s}))}  - \varepsilon \sum\limits_{i=1}^{{n_t}} {{{\tilde y_i}^t}\log (p({x_{i}^t}))} ) 
\label{eqn:ls}
\end{align}
where   $y_s$ is the labels of the source domain and ${\tilde y}^t$ is the pseudo label of target domain, obtained via Equation \ref{eq.threshold}.
It is unavoidable that a small number of pseudo labels are wrong. Thus, we add a weight parameter $\varepsilon$ to balance the impact of pseudo labels.
 
\subsection{Cross-domain Semantic Reconstruction}
     
As showing in Figure \ref{fig.framework}, our model consists of two independent generators, denoted $G^s$ and $G^t$, aiming at reducing the domain disparity in the common space. Specifically, $G^s$ is responsible for reconstructing source domain images from the common space $u$ while $G^t$ reconstructs target domain images also from $u$. The intuitive reconstruction directions are formulated as follows:
\begin{align}
 \begin{split}
    {{\tilde x}^s} &= {G^s}({u^s})  \\
    {{\tilde x}^t} &= {G^t}({u^t})  
 \end{split}   
\end{align}
where $u=E({x})$ denotes the common space feature.
To reconstruct vivid images, we use the $l_1$ pixel-wise loss to constrain the original images and reproduced images, as follows:
\begin{align}
{L_1} = \mathop {\min }\limits_{{W^E},{W^{G}}} \sum\limits_{{x^s} \in {D^s}} {\left\| {{x^s} - {{\tilde x}^s}} \right\|}  + \sum\limits_{{x^t} \in {D^t}} {\left\| {{x^t} - {{\tilde x}^t}} \right\|}
 \end{align} 
where $W^{G}$ denotes the parameters of the two generators. Note that the two generators do not share parameters, and $W^G$ is a \emph{notational convenience}.

However, the two reconstruction directions alone cannot benefit the domain information transfer, and it is necessary to build some cross-domain relationships between the two domains. Thus, we let each generator generate the other domain's data, as follows:
\begin{align}
 \begin{split} 
{{\tilde x}^{ts}} &= {G^s}({u^t})\\
{{\tilde x}^{st}} &= {G^t}({u^s})
 \end{split}   
\end{align}

Intuitively, if the pairs of ${{\tilde x}^{st}}$ and ${\tilde x}^t$, ${{\tilde x}^{ts}}$ and ${\tilde x}^s $ are reconstructed well, the learned common space adapts well on both domains. To illustrate this, taking the pair of ${{\tilde x}^{st}}$ and ${\tilde x}^t$ as an example, when they are indistinguishable, the domain disparity on the common space is small.  
On the other hand, if we remove the generator $G^s$,  it will render that the target domain space is a subspace of the source domain because of asymmetric training with only one reconstructing path \cite{DBLP:journals/corr/GhifaryKZBL16}. In the training stage, both generators are optimized adversarially until they find the best common space suitable for both domains.
   
\subsubsection{Semantic Discriminators}

It is worth noting that our model not only reconstructs the opponent domain's images but also ensures that the reproduced image has the same class label with the original input image. Specifically, following previous work \cite{Hu_2018_CVPR}, our discriminators are designed such that it distinguishes the fake and the real, and at the same time predicts the label for real images. Thus, the output of a discriminator has $N+1$ distinct values, of which $N$ values describe the image's labels and the last value defines whether the image is reconstructed or original. As in GAN \cite{NIPS2014_5423}, we can treat the GAN training stage as the generators and discriminators playing a minimax game, where the generators try to fool the discriminators by generating realistic data while the discriminators try to distinguish whether the input data is original or reconstructed. The original GAN loss is formulated as:
\begin{align}
    {L_{a}} = \mathop {\min }\limits_{W^{G}} \mathop {\max }\limits_{W^{D}} (\log (D(x)) + \log (1 - D(\tilde x)))
\end{align}
where $W^{D}$ denotes the parameters of the two (separate) discriminators and $W^G$ is the parameters of the two generators.

In our work, the discriminators need to not only differentiate fake data but also recognize the label of the real data. Thus, we augment the adversarial loss as follows:
\begin{align}
 \begin{split}
{L_{a}} = \mathop {\min }\limits_{W^{G}} \mathop {\max }\limits_{W^{D}} (&{y^s}\log (D({x^s})) + {{\tilde y}^t}\log (D({x^t})) +\\
 &{y^f}\log (D({{\tilde x}^{st}})) + {y^f}\log (D({{\tilde x}^{ts}})))
\end{split} 
\end{align}
where $y^f$ denotes the fake label and $W^{G}$/$W^{D}$ are the parameters of the generators/discriminators.
    
\subsection{Overall Objective Function}

In summary, the overall loss function is rewritten as follows:
\begin{align}
L = \mathop {\min }\limits_{{W^E},{W^{G}},{W^{D}}} ({L_c} + {L_{a}} + \alpha {L_h} + \beta {L_s} + \chi {L_1}) 
\label{all_eq}
\end{align}
where  $\alpha$, $\beta$ and $\chi$ are balance weights. 
We use  stochastic gradient descent (SGD) to  optimize the parameters. The update rules are formulated as follows, where $\eta$ is the learning rate:

\begin{align}
\begin{split}
{W^E} \leftarrow{}& {W^E} - \eta\times (\frac{{\partial {L_c}}}{{\partial {W^E}}} + \frac{{\partial {L_{a}}}}{{\partial {W^{G}}}} \times \frac{{\partial {W^{G}}}}{{\partial {W^E}}} + \alpha \frac{{\partial {L_s}}}{{\partial {W^E}}} \\
 &\qquad\qquad + \beta \frac{{\partial {L_h}}}{{\partial {W^E}}} + \chi \frac{{\partial {L_1}}}{{\partial {W^{G}}}} \times \frac{{\partial {W^{G}}}}{{\partial {W^E}}}) 
\end{split}\\
{W^{G}} \leftarrow{}& {W^{G}} - \eta\times (\frac{{\partial {L_{a}}}}{{\partial {W^{G}}}} + \chi \frac{{\partial {L_1}}}{{\partial {W^{G}}}}) \\
{W^{D}} \leftarrow{}& {W^{D}} - \eta\times (\frac{{\partial {L_{a}}}}{{\partial {W^{D}}}})  
\end{align} 

As $L_s$ is calculated by mini-batches, it is reasonable that the larger the mini-batch size is, the more accurate cluster centroids can be obtained. Moreover, the batch size must be larger than the number of classes ($N$). In our experiment, we set the batch size as $10$ times bigger of the number of labels.
  
\subsection{Differences from  Previous  Works}

\subsubsection{Difference from Duplex} 

Both Duplex~\cite{Hu_2018_CVPR} and our framework adapt the generative adversarial networks to learn the common space to reduce domain disparity. However,  Duplex uses only one generator to reconstruct cross-domain images, which requires it to add extra condition information to the common space, making it hard to align the two domains. In this work, on the other hand, we employ two separate generators, both are able to reconstruct cross-domain images. Moreover, we do not require any additional information for the common space, which allows the latent representations on the two domains to have consistent distributions.

\subsubsection{Difference from MSTN and  Co-GAN} 

Both of MSTN ~\cite{DBLP:conf/icml/XieZCC18}  and Co-GAN~\cite{DBLP:conf/nips/LiuT16} use discriminators to judge the origin of domain representations. However, the discriminators do not recognize the label of images. In contrast, our discriminators can predict class labels of images, hence are able to preserve semantic information in the common feature space, which makes the learned space more discriminative. 

\subsubsection{Difference from GTA} 

Although both GTA~\cite{DBLP:conf/cvpr/Sankaranarayanan18a} and our framework adopts the generative adversarial network to reconstruct images from the common space, GTA only uses one direction to reconstruct images: only from target to source. In contrast, our framework employs the cross-domain reconstruction strategy (both source to target and target to source), which avoids the situation that one subspace became a joint subspace affine to the opposed domain.

\section{Experiments}

We evaluate our model on the tasks of object recognition and image retrieval. Specifically, the experiments are conducted to answer the following research questions:

\textbf{RQ1:} Is our method superior to the state-of-the-art domain adaptive methods?

\textbf{RQ2:} How does each part of our model affect the performance of object recognition and image retrieval?

\textbf{RQ3:}  How well does our method learn the common space representation? 
 
\subsection{Datasets}\label{sec:datasets}

We test our method on three public digits datasets: MNIST~(M) \cite{lecun1998gradient}, SVHN (S) \cite{netzer2011reading}, and USPS (U) \cite{NIPS1988_107}. Specifically, the three datasets have the same $10$ categories but have various types. MNIST contains of 60,000 training 
and 10,000 testing images, and USPS contains 7,291 training
and 2,007 testing images. SVHN, obtained from house numbers in Google Street View images, has 73,257 training and 26,032 testing images. 

Additionally, we also evaluate our method on the standard benchmark dataset Office-31 \cite{saenko2010adapting}, which has three types of images: Amazon (A), Webcam (W) and DSLR (D). Office-31 in total has 4,110 images, of which 2,817 in Amazon, 795 in Webcam, and 498 in DSLR. 

For the digits datasets, we select three directions of domain shift: SVHN $\to$ MNIST, MNIST $\to$ USPS, and USPS $\to$ MNIST. We choose four directions of domain shifts for the Office-31 dataset: Amazon $\to$ Webcam, Webcam $\to$ Amazon, Amazon $\to$ DSLR, and DSLR $\to$ Amazon. We discard the directions of Webcam $\to$ DSLR and DSLR $\to$ Webcam because the images are highly similar. All the dataset setting are the same as  \cite{Hu_2018_CVPR}.

\subsection{Implementation Details}

The digits datasets contain simple images with the tiny size of $32 \times 32$.  For SVHN $\to$ MNIST, we use the network \cite{Hu_2018_CVPR}, which is consisted of four convolutional layers without fully connected layers. Similarly, the generator network is a symmetrical network also consisting of 4 deconvolutional layers \footnote{Our  code is released at \url{https://github.com/htlsn/igan}.}.   As for USPS $\to$ MNIST and MNIST $\to$ USPS, we use the LeNet\cite{lecun1998gradient}.  
The Office-31 dataset contains more complex images that are harder to reconstruct. Thus we choose Alexnet as the encoder network and reconstruct the $fc6$ feature layer instead of reconstructing the original images. In the training stage of Office-31, we use the model pre-trained on ImageNet to initialize our model. As for the digits datasets, we divide the training procedure into several stages. Specifically, we first use the source domain to pre-train the encoder network and then use it to initialize the first stage of the encoder network. 
After that, every several training epochs we use the last-stage encoder's parameters to initialize the next-stage encoder network, while other components are initialized with random parameters. With regards to training semantic centroid alignment, we do not precisely calculate the centers of all training samples, which is time-consuming. Instead, we calculate the centers of mini-batches to approximate the global centers. We set the batch size as $200$ for SVHN $\to$ MNIST and $100$ for MNIST $\to$ USPS and USPS $\to$ MNIST.  Plus, the length of the hash code is set as $64$ bit.

\subsection{Comparison with the State-of-the-art Domain Adaptive Hashing Methods (RQ1)}

In this section, we will compare our method with state-of-the-art works on both tasks: object recognition and image retrieval.

\subsubsection{Object Recognition}

we evaluate our method on both datasets: digits and Office-31. For the Office-31, we compare with Duplex \cite{Hu_2018_CVPR},  MSTN\cite{DBLP:conf/icml/XieZCC18},   DRCN \cite{DBLP:journals/corr/GhifaryKZBL16}, DAN \cite{DBLP:conf/icml/LongC0J15}, CDAN \cite{DBLP:conf/nips/LongC0J18}, I2I \cite{DBLP:conf/cvpr/MurezKKRK18}, and ADDA \cite{DBLP:conf/cvpr/TzengHSD17}. With regard to digits datasets, we add other two methods: Co-GAN \cite{DBLP:conf/nips/LiuT16} and CyCADA \cite{DBLP:conf/icml/HoffmanTPZISED18}. 

Table \ref{tab.digital} shows the object recognition results on the digits dataset. As can be seen, our method is superior to all the other methods on average accuracy. For the domain shift of S$\to$M, our result outperforms the current best method, Duplex, by $3.1$\%. Similarly, our result on M$\to$U is also the best, $0.3$\% higher than CDAN-M. As for U$\to$M, our result is $1.0$\% lower than Duplex. Overall, the average performance of our model is the highest and surpasses the current best model Duplex by $0.93$\%. 
Table \ref{tab.office} shows the object recognition results of the Office-31 dataset. For A $\to$ W, our method is lower by $1$\% than MSTN, but on the other three domain shifts our method outperforms the best method MSTN by $2.7$\%, $2.2$\% and $2.0$\%. Plus, the average performance is also the highest.

\subsubsection{Image retrieval}

We also test our method on the task of image retrieval. We compare our method with other unsupervised  {domain adaptive} hash methods: ITQ \cite{DBLP:journals/pami/GongLGP13}, BA \cite{DBLP:conf/cvpr/Carreira-Perpinan15}, DAH \cite{DBLP:conf/cvpr/VenkateswaraECP17}, and BDNN \cite{DBLP:conf/eccv/DoDC16}. In addition, we also test our method against the performance upper bound Supervised Hashing \cite{Li_2018_CVPR}, denoted SuH. Table \ref{tab.map} shows the image retrieval results on the Office-31 dataset, in terms of mean average precision (MAP). As stated in Section \ref{sec:datasets}, we choose three pairs of domains for image retrieval: Amazon$to$Webcam , Webcam$\to$Amazon and Amazon$\to$Dslr, following \cite{DBLP:conf/cvpr/VenkateswaraECP17}. As can be seen from Table \ref{tab.map}, for every retrieval pair, our method is superior to the other unsupervised hashing methods, with an improvement of about $3.0$\%, $2.1$\%, and $2.5$\% respectively. However, in comparison to SuH, there is still a huge gap between the supervised hash method and the supervised method. 

\begin{table}
\centering
\small
\begin{tabular}{lcccccc}\toprule
Methods & S$\rightarrow$M & M$\rightarrow$U & U$\rightarrow$M & Avg \\\midrule
CyCADA & 90.4 & 95.6 &  96.5 & 94.17 \\
Duplex & 92.5 & 96.0 & \textbf{98.8} & 95.80 \\
MSTN & 91.7 & 92.9 & - & 92.30 \\
CDAN-M & 89.2 & 96.5 & 97.1 & 94.30 \\
ADDA &  76.0 & 89.4 & 90.1& 85.17\\
Co-GAN &  - & 91.2 & 89.1 & 90.15\\
DRCN & 82.0 & 91.8 & 73.7 & 82.50\\\midrule
Ours &  \textbf{95.6} & \textbf{96.8} & 97.8 & \textbf{96.73}\\\bottomrule
\end{tabular}
\caption{Object recognition accuracies (\%) on the digits datasets.}
\label{tab.digital}
\end{table}

\begin{table}
\centering
\small
\begin{tabular}{lcccccc}\toprule
Methods & A$\rightarrow$W & A$\rightarrow$D & W$\rightarrow$A & D$\rightarrow$A & Avg\\\midrule
Duplex  & 73.2 & 74.1 & 59.1 & 61.5 & 66.96 \\
MSTN & \textbf{80.5} & 74.5 & 60.0 & 62.5& 69.38\\
I2I &  75.3 & 71.1 & 52.1 & 50.1 & 62.15 \\
CDAN-M & 78.3  & 76.3 & 57.3&  57.3 & 67.30 \\
ADDA & 73.5  & 71.6 & 53.5 & 54.6 & 63.30 \\\midrule
Ours & 79.5  & \textbf{77.2}& \textbf{62.2} &\textbf{64.5} & \textbf{70.85}\\\bottomrule
\end{tabular}
\caption{Object recognition accuracies (\%) on the Office-31 datasets.}
\label{tab.office}
\end{table}
 
\begin{table}[!ht]
\centering
\small
\begin{tabular}{lcccc}\toprule
Methods & W$\to$A & A$\to$W & A$\to$D & Avg\\\midrule
NoDA & 32.4&51.1& 51.2& 44.90\\
ITQ & 46.5& 65.2&64.3 &58.67 \\
BDNN & 49.1&65.6 &66.8 &60.50 \\
BA & 36.7& 48.0& 49.7& 44.80\\
DAH & 58.2& 71.7& 72.1 & 67.33\\\midrule
Ours & \textbf{61.2}& \textbf{73.8} & \textbf{74.6} & \textbf{69.87}\\\midrule
SuH & 88.1 & 91.6& 92.7 & 90.80 \\\bottomrule
\end{tabular}
\caption{MAP (mean average precision) 64 bits(\%) on the Office-31 datasets and Office home. Note that SuH is a supervised hashing method that represents the performance upper bound.}
\label{tab.map}
\end{table}
 
 \subsection{Ablation Study (RQ2)}
 
As presented in the previous section, our method consists of several loss functions: label prediction loss $L_c$, hash loss $L_h$, semantic centroid alignment loss $L_s$, adversarial loss $L_{a}$, and pixel-wise reconstruction loss $L_1$. In this section, we will test the effect of each component on the performance.

Table \ref{tab.ablation} shows the results of object recognition on the digits datasets, where the above loss functions are added one at a time. With only the $L_c$ component in our baseline model, the average accuracy is $69.17$. Next, we add the centroid alignment loss $L_s$. Obviously, the centroid alignment loss has a positive effect on the results with about $17.33$ improvement. Similarly, adding the hashing loss $L_{h}$ results in a further improvement of  $0.9$ over the centroid alignment loss. 

With the addition of the adversarial loss $L_{a}$ produced by our intersectant generators, the performance is  improved by $8.47$, which demonstrates that the proposed cross-domain reconstruction loss is effective.

Finally, we test  the pixel-wise loss $L_1$. As can be seen from Table \ref{tab.ablation}, $L_1$ loss further improves the performance by $0.86$. The possible reason for $L_1$ is that it can constrain the reconstructed images to look similar to the original images, which helps improve the judgement of the discriminators.

Briefly, from Table \ref{tab.ablation}, we can obtain the following two conclusions.
(1) Each of the loss function contributes positively to recognition performance, and their combination achieves the best results.
(2) The adversarial component $L_{a}$ is the key part of our framework. Specifically, when $L_{a}$ is added, about {$8.47$} improvement is obtained compared without it. Thus, the intersectant reconstruction has a significant effect on reducing domain disparity. 
 
\begin{table}
\centering
\small
\begin{tabular}{lcccccc}\toprule
Methods & S$\rightarrow$M & M$\rightarrow$U & U$\rightarrow$M & Avg \\\midrule
$L_c$ & 60.2 & 85.5 & 61.8 & 69.17 \\
$L_c$+$L_s$ & 81.2 & 89.5 & 88.8 &86.50 \\
$L_c$+$L_s$+$L_{h}$ &82.4 & 90.3  & 89.5 & 87.40 \\
$L_c$+$L_s$+$L_h$+$L_{a}$ & 94.4 & 96.1&  97.2& 95.87\\
$L_c$+$L_s$+$L_h$+$L_{a}$+$L_1$ & 95.6  & 96.8 & 97.8 &96.73 \\\bottomrule
\end{tabular}
\caption{Ablation study results of object recognition accuracy on the digits datasets.}
\label{tab.ablation}
\end{table}

\subsection{Common Space Feature Visualization (RQ3)}

\begin{figure}
\centering
\subfigure[\small{Amazon$\rightarrow$Webcam before adaptive learning.}]{
\includegraphics[width=0.45\linewidth, ]{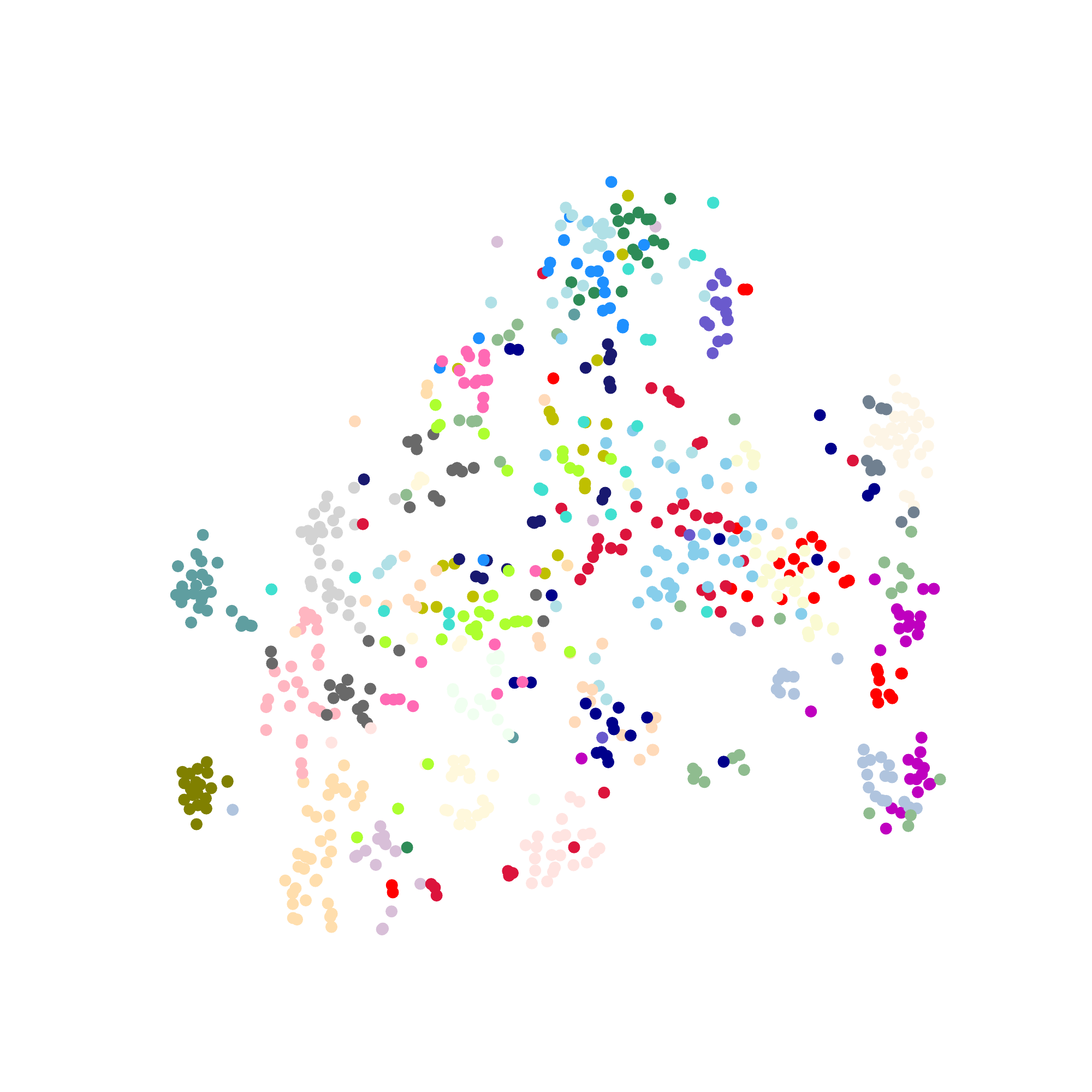}}
\label{fig.a}
\hspace{0.4cm}
\subfigure[\small{Amazon$\rightarrow$Webcam after trained by our method.}]{
\includegraphics[width=0.45\linewidth, ]{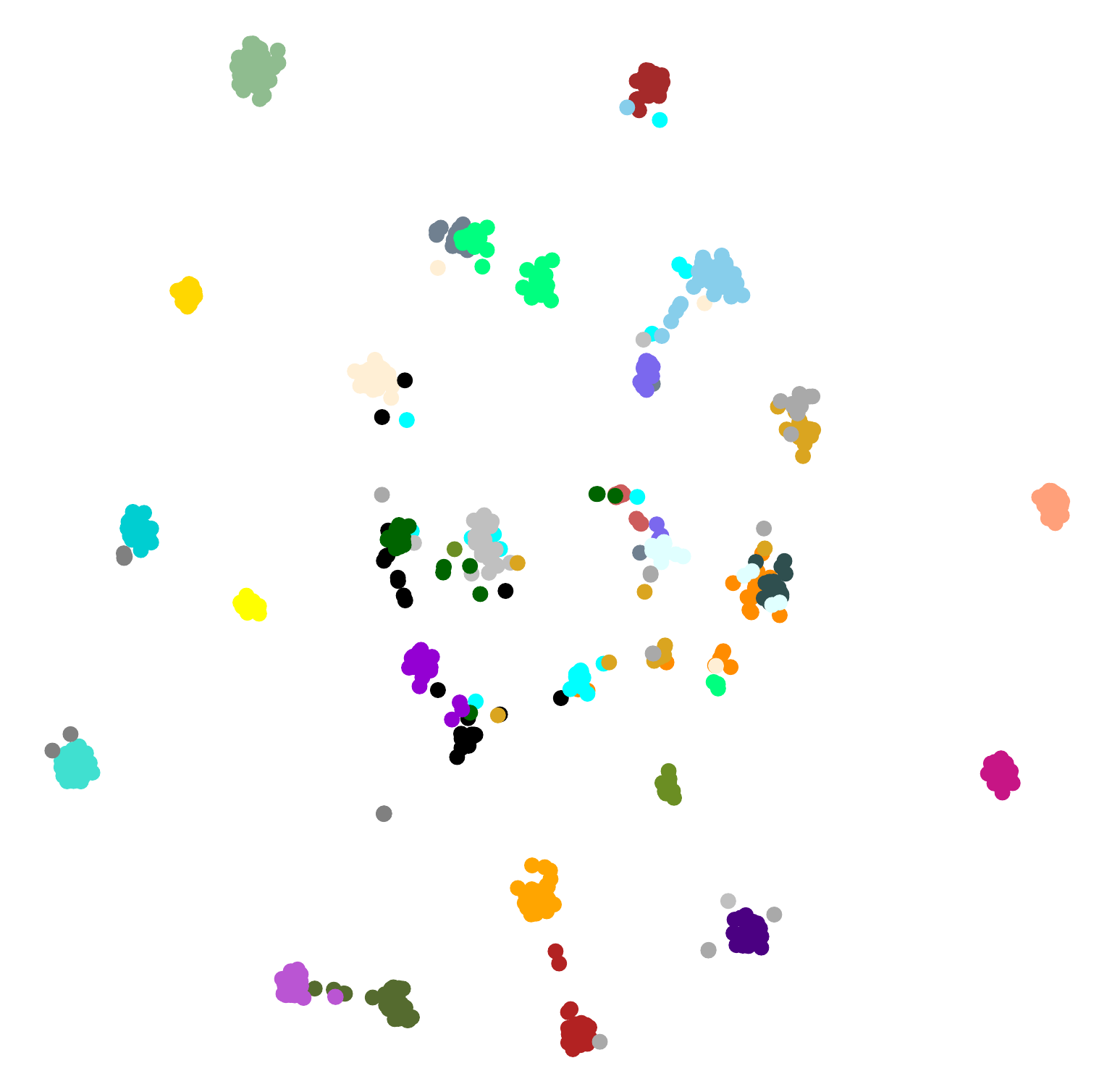}}
\label{fig.b}
\subfigure[\small{Webcam$\rightarrow$Amazon before adaptive learning.}]{
\includegraphics[width=0.45\linewidth, ]{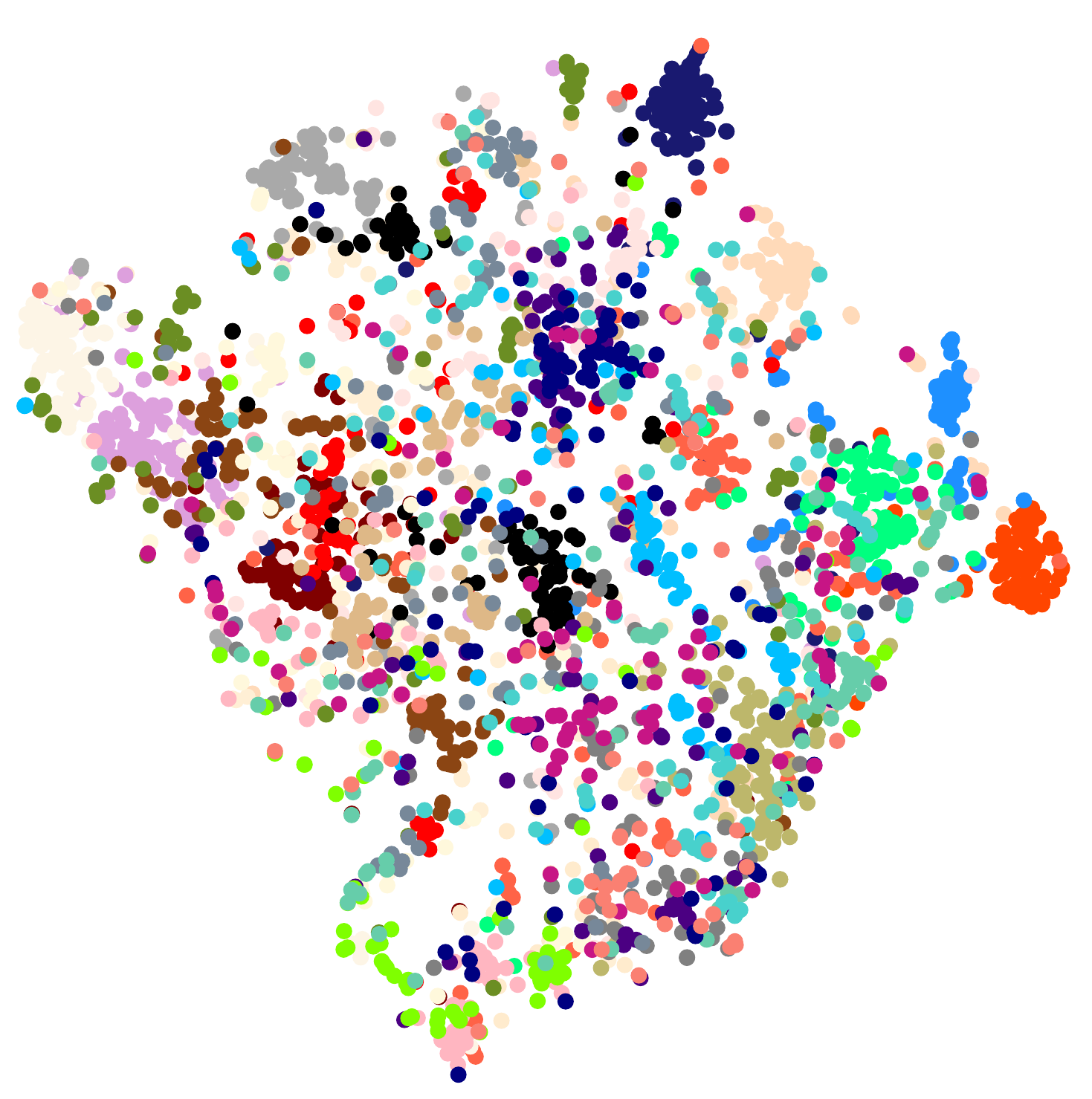}}
\label{fig.c}
\hspace{0.4cm}
\subfigure[\small{Webcam$\rightarrow$Amazon after trained by our method.}]{
\label{fig.d}
\includegraphics[width=0.45\linewidth, ]{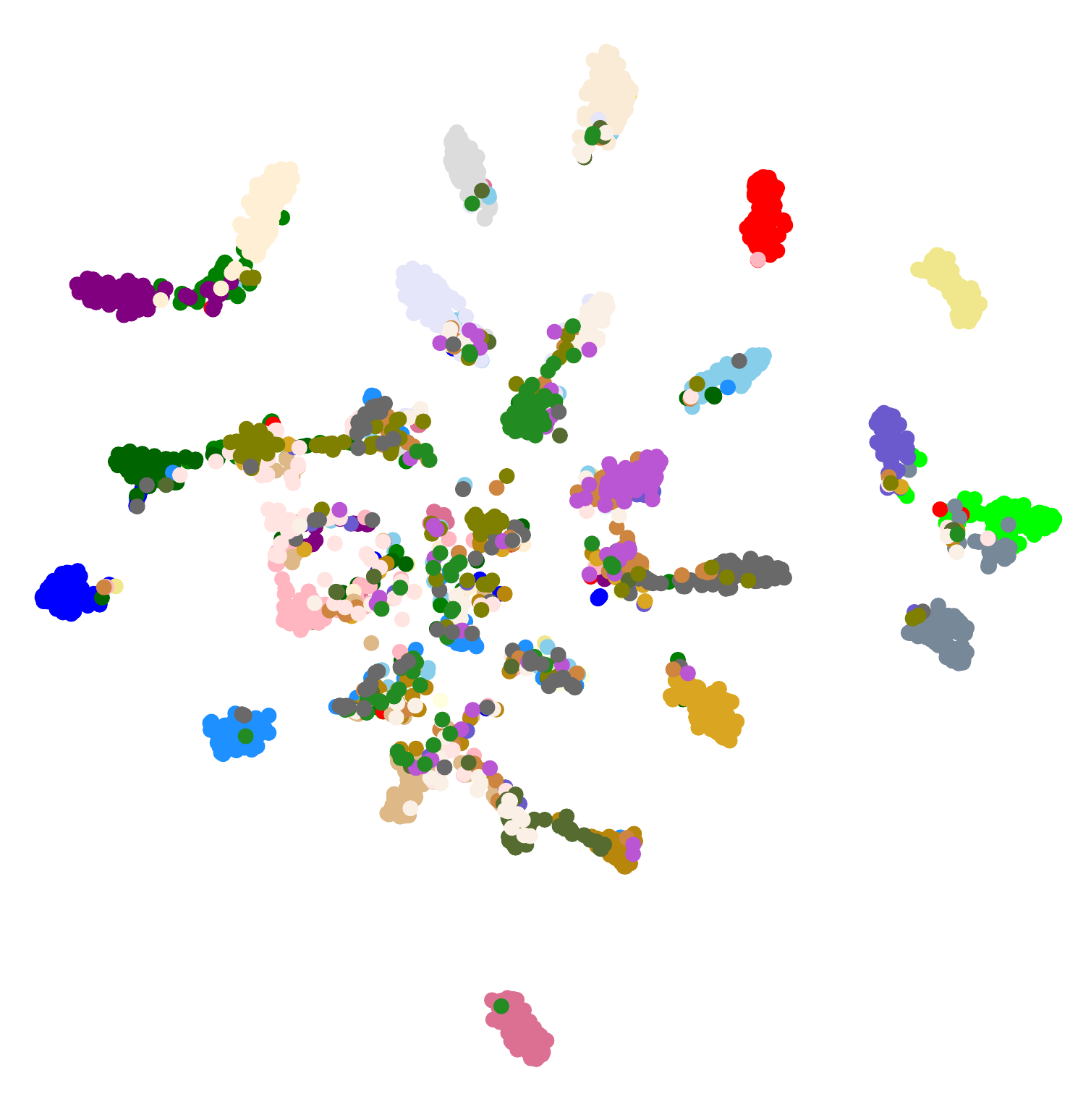}}
\caption{A visualization of the target domain features learned with and without domain adaption, with feature dimensions reduced by t-SNE from 64 to 2. For Amazon$\to$Webcam, ($a$) is the feature representation before adaptive learning, and ($b$) is trained by our method. Similarly, ($c$) is before adaptive learning and ($d$) is trained by our method, for Webcam$\to$Amazon.}
\label{fig.feat_vis}
\end{figure}

We resort to visualization to assess the quality of the common space representation learned by our method. Figure \ref{fig.feat_vis} shows the visualization of the feature representation learned by our method, with feature dimensions reduced by t-SNE from $64$ to $2$. Figure \ref{fig.feat_vis} (a) shows all the 795 points of Webcam, color-coded by the $31$ classes, that are generated by only training on the Amazon source domain. Before training by our method, it is easy to see that the points are scattered everywhere, and the points belonging to the same class (same color) do not cluster around a centroid. In contrast, after we add several loss functions \ref{all_eq}, the learned features become discriminative. I.e., points of the same class cluster in the same centroid, and class centroids have long distance, as can be seen from Figure \ref{fig.feat_vis} (b). The same observation can be made for the Webcam$\to$Amazon direction as well, as shown in Figure \ref{fig.feat_vis} (c) and (d).

\section{Conclusion}

Cross-domain hashing allows hash functions learned on one domain to be applied effectively to a new domain without supervision. In this paper, we propose an end-to-end cross-domain hashing framework based on intersectant generative adversarial networks. Our framework learns representations for both domains in a common space and combines five complementary loss functions: label prediction loss, supervised hash loss, semantic centroid alignment loss, cross-domain reconstruction loss, and pixel-wise loss. All of them help improve learning the common space and reduce the disparity between the two domains. Finally, several comparison experiments show that our method is superior to the other state-of-the-art methods in cross-domain hashing on a number of benchmark datasets.  However, compared with the supervised hashing method, there is still existing a  gap with SuH. In the future, we will focus on how to bridge this gap.  

\bibliographystyle{named}

\end{document}